\documentclass[12pt]{article}

\usepackage{sbc-template}

\usepackage{graphicx,url}
\usepackage[numbers]{natbib}
\usepackage[utf8]{inputenc}  

\usepackage{algorithm, algpseudocode}

\algnewcommand{\algorithmicforeach}{\textbf{for each}}
\algdef{SE}[FOR]{ForEach}{EndForEach}[1]
  {\algorithmicforeach\ #1\ \algorithmicdo}
  {\algorithmicend\ \algorithmicforeach}
  
\usepackage{bm,cite,array,url}

\usepackage{mathtools}
\usepackage{tablefootnote}

\newcolumntype{M}{>{\begin{varwidth}{4cm}}l<{\end{varwidth}}} 
\bmdefine\bmu{\mu}
\bmdefine\bsigma{\sigma}
\newcommand{\tab}{\hspace*{0.9em}}
\newcommand{\BigO}[1]{\ensuremath{\operatorname{O}\bigl(#1\bigr)}}
\bmdefine\bLambda{\Lambda}
\bmdefine\bSigma{\Sigma}
\graphicspath{{img/}}
\DeclareGraphicsExtensions{.pdf,.jpeg,.png,.jpg}

\title{Fast Reinforcement Learning with Incremental Gaussian Mixture Models}

\author{Rafael C. Pinto\inst{1}}

\address{Instituto Federal de Educação, Ciência e Tecnologia do Rio Grande do Sul (IFRS)\\
  Canoas, RS -- Brazil
  \email{rafael.pinto@canoas.ifrs.edu.br}
}

\begin{document} 

\maketitle

\begin{abstract}
This work presents a novel algorithm that integrates a data-efficient function approximator with reinforcement learning in continuous state spaces. An online and incremental algorithm capable of learning from a single pass through data, called Incremental Gaussian Mixture Network (IGMN), was employed as a sample-efficient function approximator for the joint state and Q-values space, all in a single model, resulting in a concise and data-efficient algorithm, i.e., a reinforcement learning algorithm that learns from very few interactions with the environment. Results are analyzed to explain the properties of the obtained algorithm, and it is observed that the use of the IGMN function approximator brings some important advantages to reinforcement learning in relation to conventional neural networks trained by gradient descent methods.
\end{abstract}
     

\section{Introduction}

Reinforcement learning is at the center of many recent accomplishments in artificial intelligence, such as playing Atari games \citep{mnih2015human} and playing \emph{go} at the grandmaster level \citep{silver2016mastering}. However, most common approaches suffer from low \emph{data efficiency}, which means that a high number of training episodes are necessary for the agent to acquire the desired level of competence on diverse tasks. The Deep Q-Learning Network (DQN) \citep{mnih2013playing} and its variants \citep{wang2015dueling,hessel2018rainbow} as well as A3C \citep{mnih2016asynchronous} and other model-free actor-critic or policy gradient methods \citep{schulman2017proximal} require millions of agent-environment interactions due to very inefficient learning. This is not acceptable for some classes of tasks, such as robotics, where failure and damage must be minimized. A robot cannot afford to fall from stairs a thousand times before learning to avoid them. More data-efficient algorithms are necessary.

Some model-based solutions to such a problem were proposed, such as \citep{gu2016continuous, kaiser2019model}, but results are still far from ideal, possibly due to its reliance on non-data-efficient function approximators like neural networks trained by gradient descent. While other works focus on the reinforcement learning side itself for data efficiency (by improving on exploration policies \citep{aubret2019survey}, for instance), the function approximation side is often neglected.

This research proposes a new solution to this problem, by integrating a sample-efficient function approximator, namely the Incremental Gaussian Mixture Network (IGMN) \citep{heinen2011igmnenia,heinen2011applying,heinen2012using}, with reinforcement learning techniques, reducing the number of required interactions with the real environment. It has been observed before that decoupling representation learning from behavior learning can be beneficial to reinforcement learning with function approximation \citep{higgins2017darla,raffin2019decoupling}, since representations can be learned much faster than behavior, which relies on sparse reward signal, while some argue that behavior can be learned faster due to slow gradient descent on representation learning \citep{blundell2016model}. IGMN can learn both representations very fast and make locally linear predictions of Q-values based on those representations, while mostly avoiding catastrophic forgetting in the process.

This work is structured as follows: section \ref{sec:related} presents related works, while section \ref{sec:igmn} presents the latest iteration of the IGMN algorithm. Section \ref{sec:figmnq} presents the proposed algorithm by combining IGMN and reinforcement learning. Section \ref{sec:experiments} shows experimental results, and section \ref{sec:conclusion} finishes this work with discussions and future works.

\section{Related Works} \label{sec:related}

In \citep{agostini2010reinforcement}, the authors use online EM to learn a single Gaussian mixture model (GMM) of the joint state-action-Q-value space. The algorithm is incremental, but it starts with some randomly placed Gaussian components instead of an empty model like the IGMN. Its component creation rule is based on an inference error threshold, as well as a Mahalanobis distance criterion. Continuous action selection is done approximately by computing the Q-value of a few random actions and selecting the one with the largest value. Matrix inverses are computed at each step, something that is avoided in the present work by using the fast variant of the IGMN algorithm (FIGMN) \citep{pinto2015fast, pinto2017scalable}. An important contribution from the authors is the use of Q-value estimation variance provided by the GMM to drive exploration, resulting in something reminiscent of Q-value sampling in Bayesian Q-learning \citep{dearden1998bayesian}. Another important insight is the need to forget old values, as Q-values are non-stationary and learned through bootstrapping, meaning that old values are mostly wrong.

In \citep{heinen2011dealing}, the IGMN was applied to a traffic simulator with continuous action selection. The algorithm is fed with the current state and the maximum stored Q-value and outputs the greedy action. Moreover, IGMN outputs the action variance too, allowing for exploration by Gaussian noise around the greedy action, with the interesting feature that this variance is also adaptive, meaning that the exploration rate adapts to each state region. IGMN can linearly generalize inside each component, making the predictions smoother. The present work is targeted at discrete action environments instead of continuous actions, and thus uses a different architecture with one Q-value per action and some additions to the learning algorithm.

Recent works on episodic memory for reinforcement learning, such as Model-Free Episodic Control (MFEC) \citep{blundell2016model} and Neural Episodic Control (NEC) \citep{pritzel2017neural}, have introduced a framework for stable, long-term, and immediately accessible memories to be stored and used analogously to Q-tables. Those models allow for generalization through k-Nearest Neighbors (kNN) search or by the use of a Radial Basis Function (RBF) layer and a linear output layer. An issue with both methods is the rapid filling of the memory, which is dealt with by removing the least recently used entries. In \citep{agostinelli2019memoryefficient}, authors propose to cluster the least recently used entry with its nearest entry, avoiding the removal of rare but important memories. That shows some resemblance to the inner workings of the IGMN algorithm, as it also stores memories as they arrive, available for immediate contribution to inference. Three main differences are: a) IGMN clusters memories continuously: there is no wait for the buffer to fill up (we observed better models constructed in this way; this feature was also added to MFEC in \citep{pinto2020model}); b) IGMN memory grows and shrinks as necessary; c) memories are complete Gaussian distributions of the joint input-output space, allowing for feature selection and linear regression inside each memory unit (this avoids the step-wise behavior observed in function approximators like kNN and RBF neural networks and requires fewer memory units). A downside of IGMN is the $\BigO{N^2}$ time complexity on the number of dimensions.

\section{Incremental Gaussian Mixture Model}\label{sec:igmn}

In the next subsections we describe the current version of the IGMN algorithm. 
	

\subsection{Learning} \label{sec:learning}
				
	The algorithm starts with no components, which are created as necessary (see subsection \ref{sec:create}). Given vector 
	$\textbf{x}$ (a single instantaneous data point), which includes both input and output concatenated, the IGMN algorithm processing step is as follows. First, the squared Mahalanobis distance $d^2_M(\textbf{x},j)$ for each component $j$ is computed:

	\begin{equation}\label{equ:figmn-maha}
		d^2_M(\textbf{x},j) =  (\textbf{x}-\bmu_j)^T \bLambda_j (\textbf{x} -\bmu_j)	\tab \forall{j}
	\end{equation}

	
\noindent	where 
	$\bmu_j$ is the $j^{th}$ component mean and $\bLambda_j$ its full precision  
	matrix (the inverse of the covariance matrix) 
	. If any $d^2_M(\textbf{x},j)$ is smaller than than $\chi^2_{D,1-\beta}$ (the $1-\beta$ percentile of a chi-squared distribution with $D$ degrees-of-freedom, where $D$ is the input dimensionality and $\beta$ is a user defined meta-parameter, e.g., $0.1$)
	, an update will occur, and posterior probabilities are calculated for each component as follows:

	\begin{equation}\label{equ:igmn-like}
		p(\textbf{x}|j) = \frac{1}{(2\pi)^{D/2}\sqrt{|\textbf{C}_j|}} \exp\left(-\frac{1}{2} d^2_M(\textbf{x},j) \right)	\tab \forall{j}
	\end{equation}

	\begin{equation}\label{equ:posterior}
		p(j|\textbf{x}) = \frac{ p(\textbf{x}|j) p(j) }{ \displaystyle\sum\limits_{k=1}^K{ 
		 p(\textbf{x}|k) p(k) } }	\tab \forall{j}
	\end{equation}	

\noindent  where $p(j)$ is the component's prior probability (equation \ref{equ:igmn-p}) and $K$ is the number of components.

After that, according to \citep{pinto2015fast, pinto2017scalable}, parameters are updated as follows. Note that rank one updates are applied directly to the precision matrices ($\bLambda$) and matrix inversions are avoided.
	
	\begin{equation}\label{equ:igmn-v}
		v_j(t) = v_j(t-1) + 1
	\end{equation}

	\begin{equation}\label{equ:igmn-sp}
		sp_j(t) = sp_j(t-1) + p(j|\textbf{x})
	\end{equation}
	
	\begin{equation}\label{equ:igmn-e}
		\textbf{e}_j = \textbf{x} - \bmu_j(t-1)
	\end{equation}
	
	\begin{equation}\label{equ:igmn-omega}
		\omega_j = \frac{ p(j|\textbf{x}) } { sp_j }
	\end{equation}

	\begin{equation}\label{equ:igmn-delta}
		\Delta\bmu_j = \omega_j \textbf{e}_j
	\end{equation}

	\begin{equation}\label{equ:igmn-mu}
		\bmu_j(t) = \bmu_j(t-1) + \Delta\bmu_j
	\end{equation}
	
	\begin{equation}\label{equ:igmn-enew}
		\textbf{e}^*_j = \textbf{x} - \bmu_j(t)
	\end{equation}

    \begin{equation}\label{equ:figmn-sherman1}
\bLambda(t) = \frac{\bLambda(t-1)}{1-\omega} + \bLambda(t-1) \textbf{e}\textbf{e}^T \bLambda(t-1)
\frac{\omega (1 - 3\omega + \omega^2)}{(\omega-1)^2 (\omega^2 - 2\omega - 1) }
    \end{equation}

		\begin{equation}\label{equ:figmn-det-lemma-apply1}
|\bSigma(t)| = (1-\omega)^D |\bSigma(t-1)| 
	    \left(1 + \frac{\omega (1 + \omega (\omega - 3)) }{1-\omega} \textbf{e}^T \bLambda(t-1) \textbf{e} \right)
	\end{equation}

	\begin{equation} \label{equ:igmn-p}
		p(j) = \frac{ sp_j } { \displaystyle\sum\limits_{q=1}^M{sp_q} }
	\end{equation}
	
\noindent	where $sp_j$ and $v_j$ are the posteriors accumulator (a probabilistic version of a counter) and the age of component $j$ (used for prunning), respectively, and $p(j)$ is its prior probability. Detailed learning steps are shown in algorithms \ref{alg:figmn-learn} and \ref{alg:figmn-update}.

\begin{algorithm}[ht]
\caption{IGMN Learning}\label{alg:figmn-learn}
\begin{algorithmic}
\Require{$\beta$, $\textbf{X}$, $\bsigma^2_{ini}$ } \algorithmiccomment{Creation threshold, dataset and initial variance vector}
\State $K = 0$, $M = \emptyset$ \algorithmiccomment{Start without any Gaussian component}
\State $\bLambda_{ini} = {({\bsigma^{2}_{ini}}\textbf{I})}^{-1}$, $|\bSigma_{ini}| = |\bLambda_{ini}|^{-1}$ \algorithmiccomment{Diagonal matrix inversion and determinant only}
\ForAll{input data vector $\textbf{x} \in \textbf{X}$}

\State $d^2_M(\textbf{x},j) =  (\textbf{x}-\bmu_{j})^T \bLambda_{j} (\textbf{x} -\bmu_{j}), \forall{j} \in M$ \algorithmiccomment{Squared Mahalanobis distance}

    \If {$\exists j$, $d^2_M(\textbf{x},j) < \chi^2_{D,1-\beta}$} \algorithmiccomment{Below $1-\beta$ percentile of a $\chi^2$ distribution}
        \State $update(\textbf{x})$ \algorithmiccomment{Algorithm \ref{alg:figmn-update}}
    \Else
        \State $K \gets K + 1$  \algorithmiccomment{Update number of components}
        \State $M \gets M \cup create(\textbf{x})$ \algorithmiccomment{Algorithm \ref{alg:figmn-create}}
    \EndIf
\EndFor
\end{algorithmic}
\end{algorithm}

\noindent where $\bsigma^2_{ini}$ represents the initial size of Gaussian components (the variance on each feature). It could be set as a fraction of the variances or sensor ranges for each dimension, if available.

\begin{algorithm}[ht]
\begin{algorithmic}
\caption{Update Components}\label{alg:figmn-update}
\Require{$\textbf{x}$} \algorithmiccomment{Input vector including known and unknown parts}
\ForAll{Gaussian component $j \in M$}

\State $d^2_M(\textbf{x},j) =  (\textbf{x}-\bmu_j)^T \bLambda_j (\textbf{x} -\bmu_j)$   \algorithmiccomment{Mahalanobis distance from input $\textbf{x}$}

\State $p(\textbf{x}|j) = \frac{1}{(2\pi)^{D/2}\sqrt{|\bSigma_j|}} \exp\left(-\frac{1}{2} d^2_M(\textbf{x},j) \right)$ \algorithmiccomment{Likelihood}

\State $p(j|\textbf{x}) = \frac{ p(\textbf{x}|j) p(j) }{ \displaystyle\sum\limits_{k=1}^K{ 
		 p(\textbf{x}|k) p(k) } }$	 \algorithmiccomment{Posterior}
\State $v_j(t) = v_j(t-1) + 1$ \algorithmiccomment{Increment age}

\State $sp_j(t) = sp_j(t-1) + p(j|\textbf{x})$ \algorithmiccomment{Accumulate posterior}
	
\State $\textbf{e}_j = \textbf{x} - \bmu_j(t-1)$    \algorithmiccomment{Compute error}
	
\State $\omega_j = \frac{ p(j|\textbf{x}) } { sp_j }$   \algorithmiccomment{Compute learning rate}

\State $\bmu_j(t) = \bmu_j(t-1) + \omega_j \textbf{e}_j$    \algorithmiccomment{Update mean}
	
\State $\bLambda(t) = \frac{\bLambda(t-1)}{1-\omega} + \bLambda(t-1) \textbf{e}\textbf{e}^T \bLambda(t-1)
\frac{\omega (1 - 3\omega + \omega^2)}{(\omega-1)^2 (\omega^2 - 2\omega - 1) }$ \algorithmiccomment{Update precision matrix}

\State $p(j) = \frac{ sp_j } { \displaystyle\sum\limits_{q=1}^M{sp_q} }$    \algorithmiccomment{Update prior}

\State $|\bSigma(t)| = (1-\omega)^D |\bSigma(t-1)| 
	    \left(1 + \frac{\omega (1 + \omega (\omega - 3)) }{1-\omega} \textbf{e}^T \bLambda(t-1) \textbf{e} \right)$ \algorithmiccomment{Update determinant}
	
\EndFor
\end{algorithmic}
\end{algorithm}

\subsection{Creating New Components} \label{sec:create}
	
	If the update condition in algorithm \ref{alg:figmn-learn} is not met, then a new component $j$ is created and initialized as show in algorithm \ref{alg:figmn-create}.

\begin{algorithm}[ht]
\caption{Create Component}\label{alg:figmn-create}
\begin{algorithmic}
\Require{$\textbf{x}$} \algorithmiccomment{Full input vector containing  concatenated inputs and outputs}
\State \Return new Gaussian component $j$ with $\bmu_j = \textbf{x}$, $\bLambda_j = \bLambda_{ini}$, $|\bSigma_j| = |\bSigma_{ini}|$, $sp_j = 1$, $v_j = 1$, $p(j) = \frac{1}{\displaystyle\sum\limits_{k=1}^K{sp_k}}$
\end{algorithmic}
\end{algorithm}
	
\subsection{Inference} \label{sec:recalling}


	In IGMN, any element can be predicted by any other element. This is done by reconstructing data from the target elements ($\textbf{x}_t$, a slice of the entire input vector $\textbf{x}$) by estimating the posterior probabilities using only the given elements ($\textbf{x}_i$, also a slice of the entire input vector $\textbf{x}$), as follows:

	\begin{equation}\label{equ:recall}
		p(j|\textbf{x}_i) = \frac{ p(\textbf{x}_i|j) p(j) }{ \displaystyle\sum\limits_{q=1}^M{ 
		 p(\textbf{x}_i|q) p(q) } }	\tab \forall{j}
	\end{equation}		
	It is similar to equation \ref{equ:posterior}, except that it uses a modified input vector $\textbf{x}_i$ with the target 	elements $\textbf{x}_t$ removed from calculations. After that, $\textbf{x}_t$ can be reconstructed using the conditional mean:
	
	\begin{equation}\label{equ:figmn-reconstructfull}
		\hat{\textbf{x}_t} = \displaystyle\sum\limits_{j=1}^M{ p(j|\textbf{x}_i) (\bmu_{j,t} - \bLambda_{j,it}\bLambda_{j,t}^{-1} (\textbf{x}_i - \bmu_{j,i})) }\,,
	\end{equation}

\noindent	where $\bLambda_{j,it}$ is the submatrix of the $j$th component's precision matrix associating the known ($i$) and unknown ($t$) 
	parts of the data, $\bLambda_{j,t}$ is the submatrix corresponding to the unknown ($t$) part only, $\bmu_{j,i}$ is the $j$th's 
	component mean's known part ($i$) and $\bmu_{j,t}$ its unknown part ($t$). This procedure can be seen in algorithm \ref{alg:figmn-inference}, while the covariance matrix block decomposition is shown in equation \ref{equ:blockdecomposition}.
	
\begin{equation}\label{equ:blockdecomposition}
\begin{split}
\bLambda_j = \begin{bmatrix}
\bSigma_{j,i} & \bSigma_{j,it} \\
\bSigma_{j,ti} & \bSigma_{j,t}
\end{bmatrix}^{-1} = 
\begin{bmatrix}
\bLambda_{j,i} & \bLambda_{j,it} \\
\bLambda_{j,ti} & \bLambda_{j,t}
\end{bmatrix} \\
= \begin{bmatrix}
(\bSigma_{j,i} - \bSigma_{j,it}\bSigma_{j,t}^{-1}\bSigma_{j,ti})^{-1} & -\bSigma_{j,i}^{-1}\bSigma_{j,it}(\bSigma_{j,t} - \bSigma_{j,ti}\bSigma_{j,i}^{-1}\bSigma_{j,it})^{-1} \\                 -\bSigma_{j,t}^{-1}\bSigma_{j,ti}(\bSigma_{j,i} - \bSigma_{j,it}\bSigma_{j,t}^{-1}\bSigma_{j,ti})^{-1} & (\bSigma_{j,t} - \bSigma_{j,ti}\bSigma_{j,i}^{-1}\bSigma_{j,it})^{-1}  
\end{bmatrix}  \,.
\end{split}
\end{equation}

\begin{algorithm}[ht]
\caption{Inference (recall)}\label{alg:figmn-inference}
\begin{algorithmic}
\Require{$\textbf{x}_i$} \algorithmiccomment{Input vector containing only the known elements}

\State $\bSigma_{j,i}^{-1} = \bLambda_{j,i} - \bLambda_{j,it}\bLambda_{j,t}^{-1}\bLambda_{j,ti}$, \tab $\forall{j} \in M$ \algorithmiccomment{Inverse of input portion}

\State $|\bSigma_{j,i}| = |\bSigma_{j}| |\bLambda_{j,t}|$, \tab $\forall{j} \in M$ \algorithmiccomment{Determinant of input portion}

\State $d^2_M(\textbf{x}_i,j) =  (\textbf{x}_i-\bmu_{j,i})^T \bSigma_{j,i}^{-1} (\textbf{x}_i -\bmu_{j,i})$, \tab $\forall{j} \in M$ \algorithmiccomment{Squared Mahalanobis distance}

\State $p(\textbf{x}_i|j) = \frac{1}{(2\pi)^{D/2}\sqrt{|\bSigma_{j,i}|}} \exp\left(-\frac{1}{2} d^2_M(\textbf{x}_i, j) \right)$, \tab $\forall{j} \in M$ \algorithmiccomment{Likelihoods}

\State $p(j|\textbf{x}_i) = \frac{ p(\textbf{x}_i|j) p(j) }{ \displaystyle\sum\limits_{k=1}^{k \in M}{ 
		 p(\textbf{x}_i|k) p(k) } }$, \tab $\forall{j} \in M$	 \algorithmiccomment{Posteriors}

\State \Return $\displaystyle\sum\limits_{j=1}^{j \in M}{ p(j|\textbf{x}_i) (\bmu_{j,t} - \bLambda_{j,it}\bLambda_{j,t}^{-1} (\textbf{x}_i - \bmu_{j,i})) }$ \algorithmiccomment{Weighted conditional mean}

\end{algorithmic}
\end{algorithm}

\noindent Note that it requires a matrix inversion of $\bLambda_{j,t}$, which is used in two different places but can be reused in order to avoid two inversions. Also, since in most applications the number of outputs is much smaller than the number of inputs, this does not impose any huge penalty in complexity. Also note that the original FIGMN paper was missing the first two formulas, but they were derived and presented in \citep{chamby2018adaptive}.

\subsection{Removing Spurious Components} \label{sec:removing}
	A component $j$ is removed whenever $v_j > v_{min}$ and $sp_j < sp_{min}$, where $v_{min}$ and $sp_{min}$ are manually chosen (e.g., 5.0 and 3.0,	respectively). In that case, also, $p(k)$ must be adjusted for all $k \in K$, $k \ne j$, using (\ref{equ:igmn-p}). In other words, each component is given some time $v_{min}$ to show its importance to the model in the form of an accumulation of its posterior probabilities $sp_j$. We do not use component pruning in this work, since it produced deleterious results when combined with reinforcement learning (rare but still important experiences may be forgotten, such as reaching the goal in the Mountain Car problem). New removal procedures that protect high reward memories must be investigated.

\section{Reinforcement Learning with IGMN}\label{sec:figmnq}

Here we develop a data-efficient (and here we define it empirically) reinforcement learning algorithm for continuous state spaces and discrete action spaces. We argue that other algorithms' inefficiencies come partially from the slow function approximators they use, i.e., neural networks trained by gradient descent. Hence, a data-efficient function approximator should be used instead. 
The algorithm chosen here is the IGMN, due to its speed and versatility. IGMN can be combined with reinforcement learning in various ways \citep{heinen2011dealing}, but here we focus on the most successful one, called Unified FIGMN-Q in \citep{pinto2017continuous}. For the sake of simplicity, it is called Q-IGMN from now on.

\begin{figure}[ht]
\centering  \includegraphics[width=0.9\textwidth]{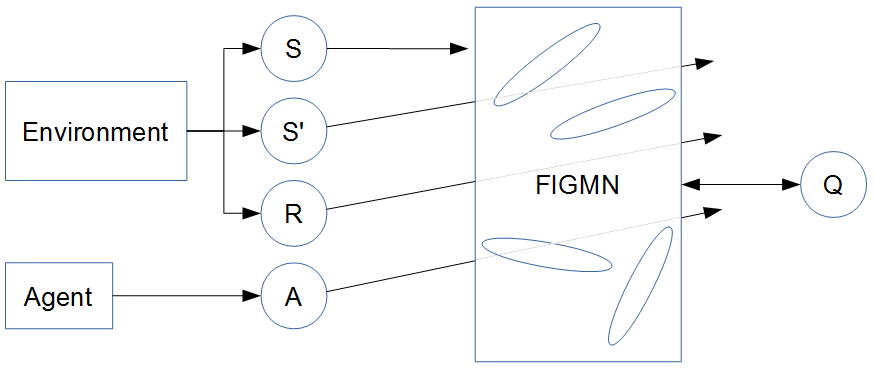}
\caption{Q-IGMN: The IGMN models the densities of the joint space of states ($S$) and Q-values for each action. Resulting states ($S'$), rewards ($R$) and actions ($A$) are used in the learning algorithm to form the Q-learning target. Q-learning is embedded into the IGMN itself as it corrects the Q-values.} \label{fig:ufigmn-q}
\end{figure}

Q-IGMN is achieved by modeling the joint space of states and Q-values (one Q-value for each possible action), as this is IGMN's standard way of doing supervised learning. This architecture is shown in figure \ref{fig:ufigmn-q}. It allows only discrete actions, but, on the other hand, allows for fast action selection (just input the current state and select the action corresponding to the largest Q-value). Algorithm \ref{alg:u-figmn-q} describes the behavior of this architecture.

\begin{algorithm}[ht]
\caption{Q-IGMN Algorithm}\label{alg:u-figmn-q}
\begin{algorithmic}
\Require{$A$ is a set of actions, 
    $D$ is the number of dimensions in the state space,
    $\gamma$ is the discount factor,
    $\epsilon$ is the exploration rate,
    $E_{max}$ is the maximum size of the experience replay stack
}
\State Initialize IGMN
\State Initialize empty experience replay stack $E$ with arbitrary maximum size $E_{max}$
\State Observe current state $\textbf{s}$   \algorithmiccomment{Initial state}
\Repeat
    \If{IGMN is empty or $U(0,1) < \epsilon$}   \algorithmiccomment{$\epsilon$-greddy policy}
        \State $a \leftarrow $ random action from $A$   \algorithmiccomment{Exploration}
    \Else
        \State $Q[\textbf{s},.] \leftarrow$ IGMN.recall($\textbf{s}$) \algorithmiccomment{obtain $Q_a$ for each action in $A$}
        \State $a \leftarrow argmax_{a'}Q[\textbf{s},a']$   \algorithmiccomment{Exploitation}
    \EndIf
    \State perform action $a$ and observe reward $r$ and state $\textbf{s}'$
    \State push $\{\textbf{s}, a, r, \textbf{s}'\}$ into $E$
    \If{$E$ is full or episode ended}   \algorithmiccomment{Learning happens here}
        \While{$E$ not empty}
            \State pop $\{\textbf{s}$, $a$, $r$, $\textbf{s}'\}$ from $E$   \algorithmiccomment{Extract last experience from stack}
            \State $Q'[\textbf{s}',.] \leftarrow$ IGMN.recall($\textbf{s}'$)    \algorithmiccomment{Compute Q' for next state}
            \State $a_{max} \leftarrow argmax_{a'}Q'[\textbf{s}',a']$   \algorithmiccomment{Greedy action for next state}
            \State $Q_a \leftarrow null, \forall{a} \in A$    \algorithmiccomment{Create vector of $nulls$}
            \State $Q_{a_{max}} \leftarrow r + \gamma Q'[\textbf{s}',a_{max}]$ \algorithmiccomment{Target computed according to greedy action}
            \State $\textbf{x} \leftarrow \{s_1, s_2, ..., s_D, Q_1, Q_2, ..., Q_{|A|}\}$ \algorithmiccomment{Concatenate $\textbf{s}$ and $Q[\textbf{s},.]$}
            \State IGMN.update($\textbf{x}$) \algorithmiccomment{$null$ values do not update IGMN parameters}
        \EndWhile
    \EndIf
    \State $\textbf{s} \leftarrow \textbf{s}'$
\Until{termination}

\end{algorithmic}
\end{algorithm}

\section{Experiments and Results} \label{sec:experiments}

OpenAI Gym \citep{OpenAIGym} has been used as the primary platform for testing the proposed algorithm and comparing it to other popular alternatives. This open-source platform provides ready-to-use reinforcement learning environments and is used to provide a leaderboard-like page\footnote{This feature has since been removed from the platform, but experiment result files are still accessible} comparing the performance of algorithms from different users, including the most common algorithms like Q-learning. The performance is measured by two factors: episodes to solve (data efficiency) and mean reward. Each environment has some goal accumulated reward to reach. More precisely, the agent must obtain an average accumulated reward equal or higher than the goal for 100 consecutive episodes, making it a very robust experiment. Time to solve is defined as the first episode of the successful 100 episode window.

More demanding tasks like the Atari games \citep{bellemare2012arcade} were avoided due to hardware restrictions, but are planned for future experiments. All experiments were executed on an Intel i7 laptop without access to GPU. 

Since the OpenAI Gym platform currently only supports the Python language, this was the language of choice for implementing the final experiments. An existing open-source implementation \footnote{https://github.com/renatopp/liac/blob/master/liac/models/igmn.py} of the IGMN algorithm in this language was used as a basis for the more scalable FIGMN algorithm implementation. 

\begin{figure}[ht]
\centering  \includegraphics[width=0.9\textwidth]{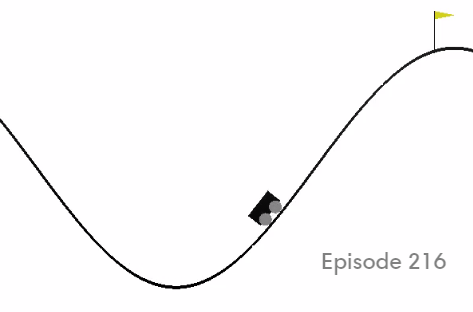}
\caption{The mountain car environment running inside OpenAI Gym.} \label{fig:mountaincar}
\end{figure}

The mountain car task consists of controlling an underpowered car to reach the top of a hill. It must go up the opposite slope to gain momentum first. The agent has three actions at its disposal, accelerating it leftwards, rightwards,
or no acceleration at all. The agent’s state is made up of two features: current position and speed. Only 200 steps are available for the agent to explore during each episode. This task is considered solved after 100 consecutive episodes with an average of 110 steps or less to reach the top of the hill. A screenshot of this environment can be seen in figure \ref{fig:mountaincar}.

\begin{figure}[ht]
\centering  \includegraphics[width=0.9\textwidth]{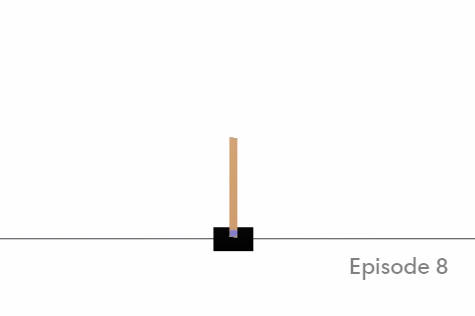}
\caption{The cart-pole environment running inside OpenAI Gym.} \label{fig:cartpole}
\end{figure}

The cart-pole task consists of balancing a pole above a small car that can move left or right at each time step. Four variables are available as observations: current position and speed of the cart and current angle and angular velocity of the pole. Version 0 requires the pole to be balanced for 200 steps, while version 1 requires 500 steps. This task is considered solved after 100 consecutive episodes with an average of 195 steps for version 0 and 475 steps for version 1 without dropping the pole.  A screenshot of this environment can be seen in figure \ref{fig:cartpole}

\begin{figure}[hbt]
\centering  \includegraphics[width=0.9\textwidth]{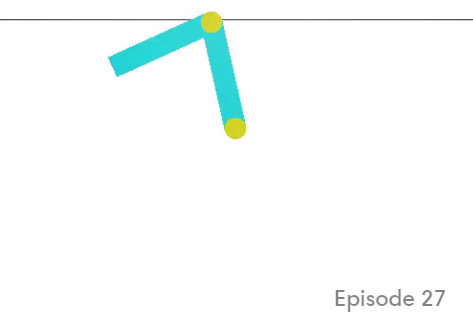}
\caption{The acrobot environment running inside OpenAI Gym.} \label{fig:acrobot}
\end{figure}

Finally, the acrobot task requires a 2-joint robot to reach a certain height with the tip of its "arm". Torque in two directions can be exerted on the 2 joints, resulting in 4 possible actions. The current angle and angular velocity of each joint are provided as observations. There are 200 steps per episode available for exploration. This task is considered solved after 100 consecutive episodes with an average of 100 or fewer steps to reach the target height. A screenshot of this environment can be seen in figure \ref{fig:acrobot}

The Q-IGMN algorithm was compared to other 3 algorithms with high scores on OpenAI Gym: Sarsa($\lambda$) with Tile Coding, Trust Region Policy Optimization \citep{schulman2015trust} and Dueling Double DQN \citep{wang2015dueling}. These algorithms were chosen according to the OpenAI Gym rank at the time of writing. Table \ref{episodes} shows the number of episodes required for each algorithm to reach the required reward threshold for the 3 tasks \footnote{The OpenAI Gym platform computes results based only on successful runs}. Results were extracted on June-2016 and January-2017.

\begin{table}[thb]
\caption{\label{episodes}Number of episodes to solve each task.}
\scriptsize
{\centering \begin{tabular}{cccccc}
\\
\hline
Environment & Q-IGMN \tablefootnote{\url{https://gym.openai.com/algorithms/alg_gt0l11Rf6hkwUXjVsRcw}}
& Sarsa($\lambda$) \tablefootnote{\url{https://gym.openai.com/algorithms/alg_hJcbHruxTLOa1zAuPkkAYw}} & TRPO \tablefootnote{\url{https://gym.openai.com/algorithms/alg_yO8abVs8Spm21Icr60SB8g}}   & Duel DDQN \tablefootnote{\url{https://gym.openai.com/algorithms/alg_zy3YHp0RTVOq6VXpocB20g}} \\
\hline
Cart-Pole V0 & \textbf{0.5} $\pm$ 0.56 & 557 & 	2103.50 $\pm$ 3542.86 & 	51.00 $\pm$ 7.24 \\
Mountain Car V0 & \textbf{0.0} & 1872.50 $\pm$ 6.04 & 4064.00 $\pm$ 246.25 & - \\
Acrobot V0* & \textbf{0.0} & 742 & 2930.67 $\pm$ 1627.26 & 31 \\
\hline
\multicolumn{6}{c}{* This task is not available on the OpenAI Gym server anymore, so the result can be verified only locally.}
\end{tabular} \scriptsize \par}
\end{table}

\begin{figure}[hbt]
\centering  \includegraphics[width=0.9\textwidth]{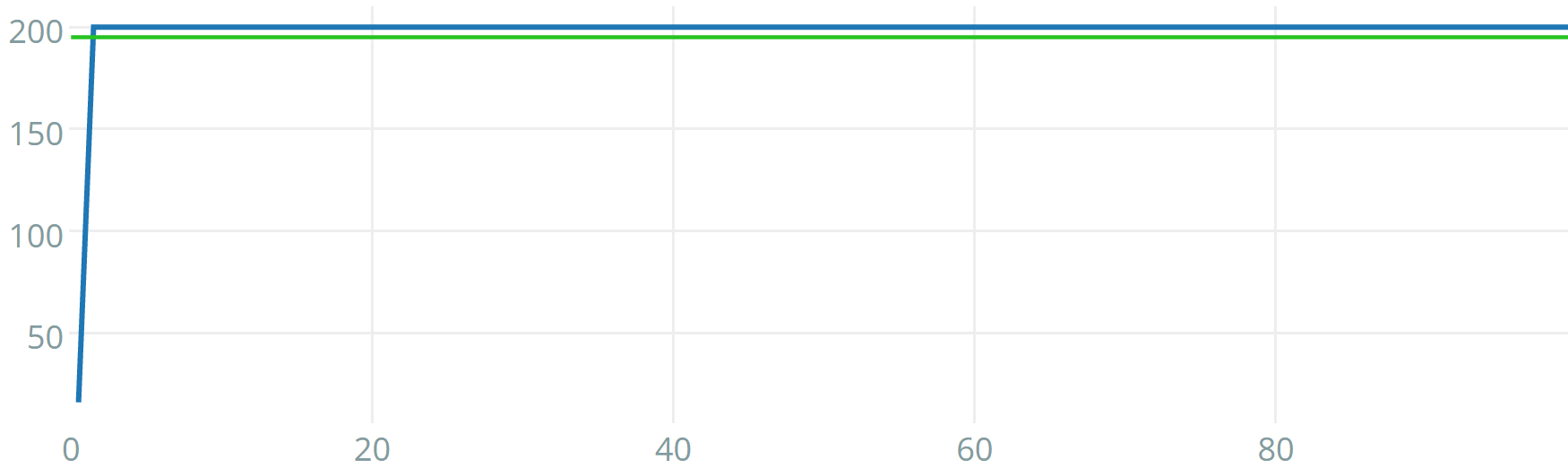}
\caption{Example of a single evaluation of the Q-IGMN algorithm on the cart-pole v0 environment.} \label{fig:ufigmnq-cartpole-v0}
\end{figure}

\begin{figure}[hbt]
\centering  \includegraphics[width=0.9\textwidth]{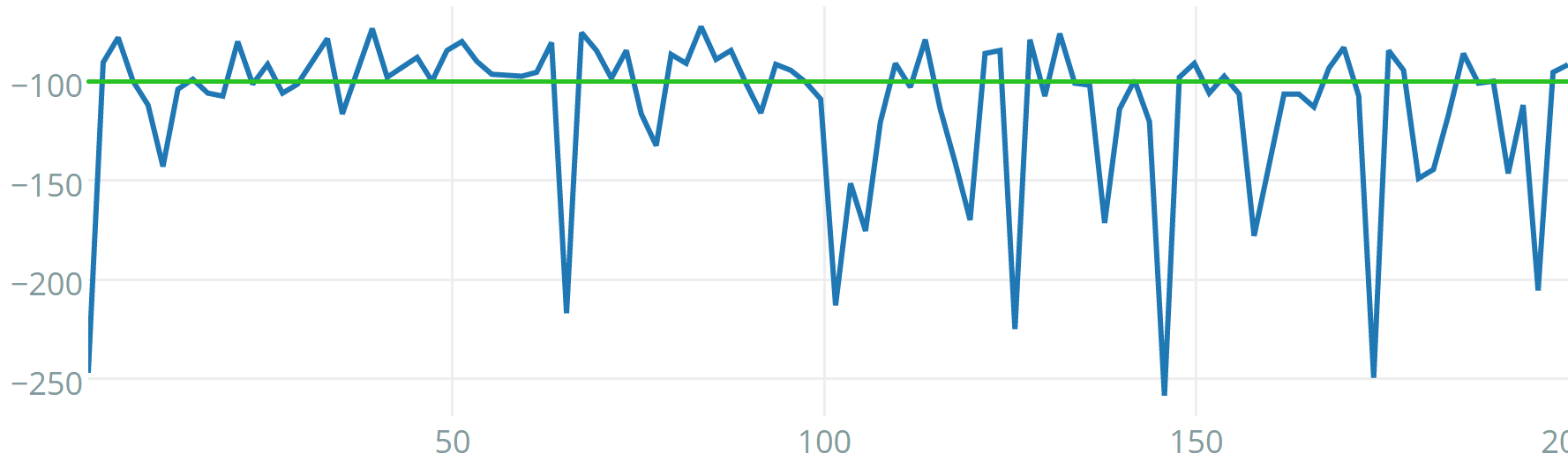}
\caption{Example of a single evaluation of the Q-IGMN algorithm on the acrobot v1 environment.} \label{fig:ufigmnq-acrobot-v1}
\end{figure}

\begin{figure}[hbt]
\centering  \includegraphics[width=0.9\textwidth]{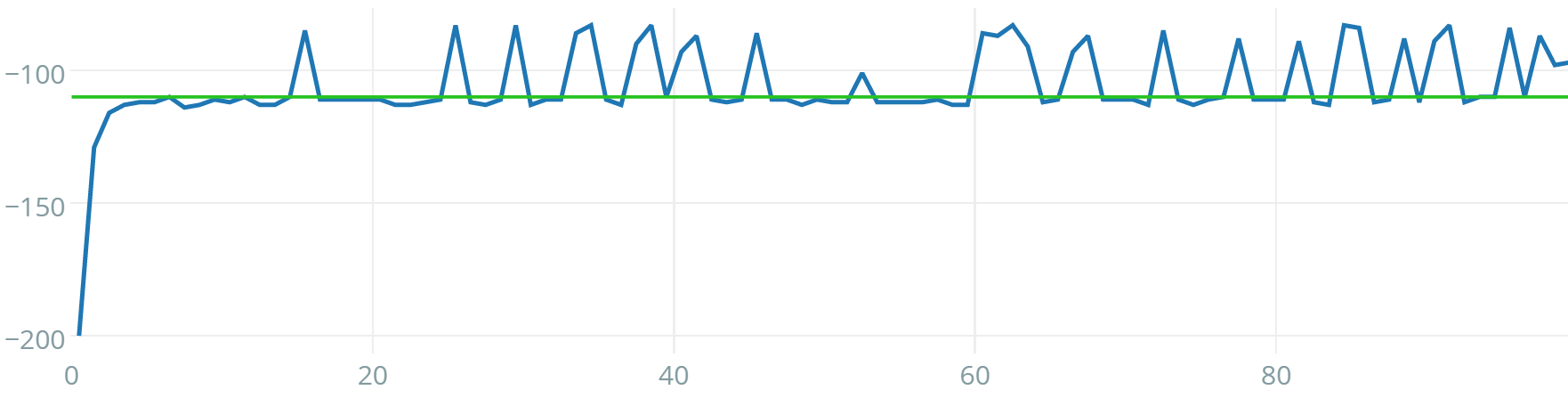}
\caption{Example of a single evaluation of the Q-IGMN algorithm on the mountain car v0 environment.} \label{fig:ufigmnq-mountaincar-v0}
\end{figure}

Albeit being very difficult to tune, Q-IGMN proved to be the most data-efficient reinforcement learning algorithm in this set of experiments. It was able to solve each of the tasks in a few episodes. Typical learning curves for this algorithm in all tasks are shown in figures \ref{fig:ufigmnq-cartpole-v0}, \ref{fig:ufigmnq-acrobot-v1} and \ref{fig:ufigmnq-mountaincar-v0}. Note that, when this experiment was performed, acrobot v0 was not available anymore at the Gym server, so experiments were kept locally. The acrobot learning curve shown is for v1, where physics where improved, there are 500 available time steps per episode instead of 200 and there is no reward threshold, so it is not trivial to compare algorithms regarding data-efficiency (hence why we use v0 in the comparison table). Its learning curve is still informative, nevertheless. Another interesting result is that Q-IGMN solved most of the tasks with a single Gaussian component, implying that their Q-value function is (at least approximately) linear. The mountain car task, on the other hand, required 8 Gaussian components (its Q-value function has a spiral surface).

 However, a single trick was essential to guarantee this data-efficiency: we employed a kind of experience replay buffer. But instead of sampling it randomly and repeatedly at each time step (as commonly done in most works with neural networks), it was sampled from the most recent observation to the oldest one (randomly sampling the experience replay buffer also works here, but performance degrades), \emph{in a single pass} (which means that we still perform only one update per step, they are just shifted and accumulated), and learning only happens when the buffer is full; after that, it is emptied again. Interestingly, time-correlated data does not impair IGMN's performance as it does with neural networks. It is shown in the original Incremental Gaussian Mixture Model paper \citep{engel2010concept} that data should vary slowly (i.e., it should not be independent and identically distributed (i.i.d.), exactly the opposite condition for neural networks). From another point-of-view, this could be seen as mini-batch learning. This technique drastically improved the algorithm, and there seem to be 2 effects taking place to explain this:

\begin{itemize}

 \item First, it is common knowledge that conventional Q-learning with function approximation diverges due to the non-stationary nature of the Q-values \citep{sutton1998reinforcement}. The Q-learning update rule uses the function approximator itself to provide a target, which changes immediately, while action selection is also done using the same ever-changing approximator. By doing updates in mini-batches, this issue is minimized, as action selection is performed over a stable function approximator. Then, it is updated all at once, and after that, actions can be selected from a stable approximator again. This bears resemblance to Double Q-Learning, where 2 estimators are used to select actions from a stable approximator which is updated once in a while from the second estimator (which updates constantly), except we use a single estimator with sporadic updates instead;
 
 \item The second effect produced by this mini-batch approach is similar to trace decays: while conventional Q-learning updates a single state-action per step, trace decays allow us to update a large portion of the state-action history at once. So, when a goal is reached, its value is rapidly propagated backward. The mini-batch approach results in something very much like it: since the most recently visited states are updated first, older states will receive updated values immediately, eliminating the need to visit those states again to "see" the new values. A key difference is that trace decays perform all these updates at \emph{every} time step, resulting in high computational demands.
    
\end{itemize}

In general, larger buffer sizes improved the results. For small episodic tasks like the ones presented here, it is enough to set the maximum buffer size as the maximum number of steps per episode for each task, resulting in episodic batch updates (note, however, that IGMN updates are always incremental, i.e., each experience is presented to the algorithm and then discarded). The resulting buffer sizes are vastly smaller than conventional experience replay buffers for deep learning.

Additionally, in the mountain car task, it was
necessary to apply other techniques to ensure stability and convergence:

\begin{itemize}

\item The first one was to set an independent learning rate $\alpha$ (with its own annealing schedule) for the Q-value variables in the Q-IGMN. It means that equation \ref{equ:igmn-omega} (reproduced below) should only apply to the state variables when updating the mean and also the precision matrix.
	\begin{equation}\label{equ:igmn-omega2}
		\omega_j = \frac{ p(j|\textbf{x}) } { sp_j }\,.
	\end{equation}
When updating the means and precision matrices for variables corresponding to the Q-values, the following equation should be used instead:
	\begin{equation}\label{equ:igmn-omega-q}
		\omega_j = p(j|\textbf{x}) \alpha\,.
	\end{equation}
This effectively decouples representation learning speed from behavior learning speed, and is necessary in some cases, since the default IGMN learning rate $\omega$ given by the original equation decreases very fast, which may not be appropriate for a given task. Also, the original equation results in the component mean converging to the true mean of all input data, which is expected when dealing with stationary data, but not with non-stationary data as is the case of the Q-values. The same problem was found by \citep{agostini2010reinforcement} and solved in a way that was appropriate for the used algorithm, but it would be analogous to applying a decay rate to $sp_j$ in Q-IGMN, thus making $\omega$ decrease slower. It is not the same as the independent learning rate, as it affects all variables and not only the Q-values. In our experiments, the independent learning rate for Q-values was able to find a solution, while $sp$ decay was not.

\item The second one was a technique akin to early stopping, which is used in neural networks to avoid overfitting. Here, a reward threshold (in this case $-122$) was set in order to stop learning. When this threshold is reached, the $\epsilon$ exploration parameter, the $\alpha$ learning rate, and the $\tau$ component creation threshold are all set to $0$, effectively stopping any learning. This is necessary to avoid new components being created in overlapping positions with old ones, producing catastrophic forgetting. An alternative, which is explored in \citep{chamby2018adaptive}, is to improve the stability of IGMN itself by avoiding overlaps automatically.

\end{itemize}

\begin{figure}[hbt]
\centering  \includegraphics[width=0.9\textwidth]{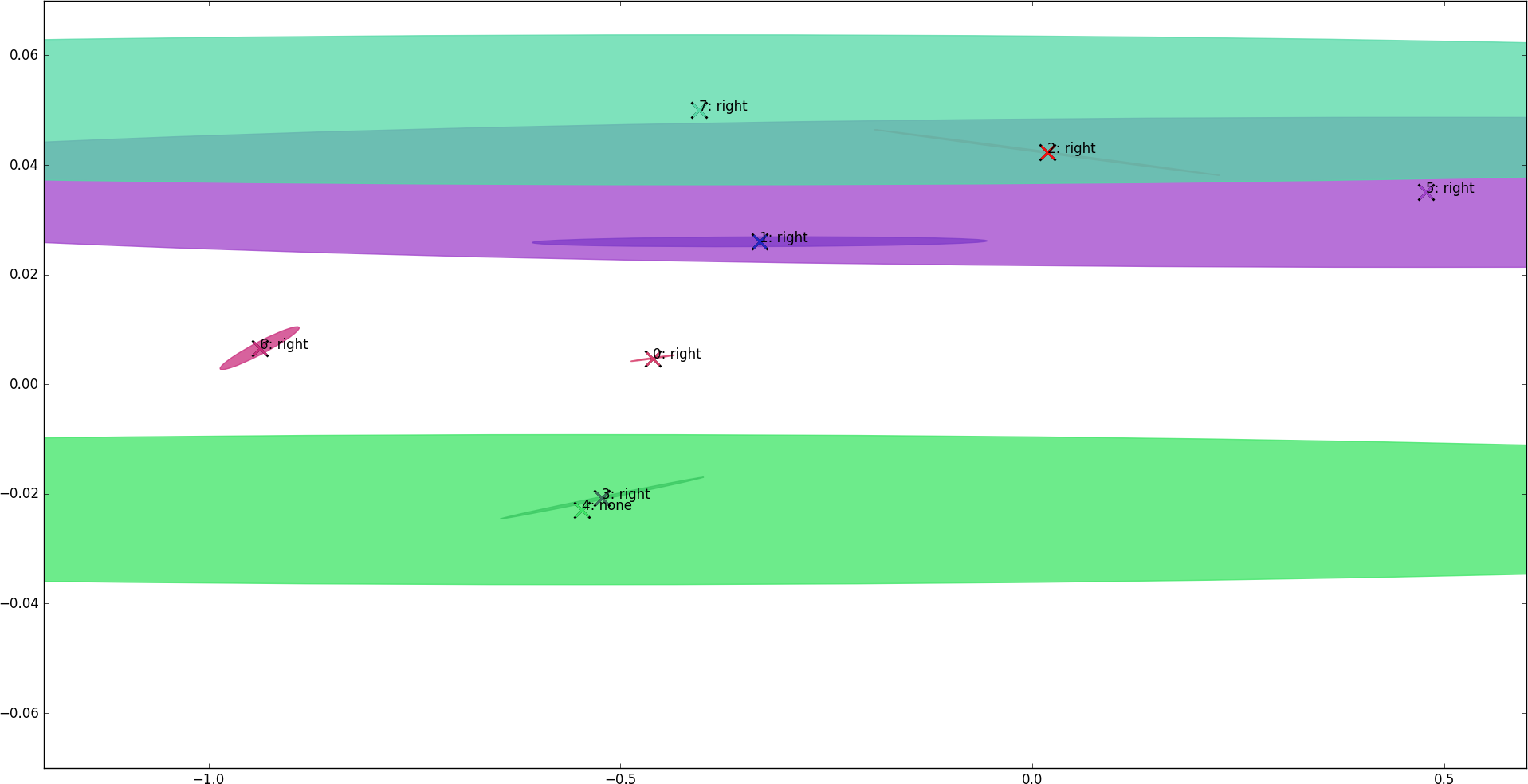}
\caption{Policy learned by the Q-IGMN algorithm on the mountain car task. The coloured bands are just very elongated ellipses.} \label{fig:ufigmnq-mountaincar-policy}
\end{figure}

The learned policy is shown in figure \ref{fig:ufigmnq-mountaincar-policy}. Actions are shown according to the largest Q-value in the Gaussian means. It is possible to verify that most components suggest the "right" action, having a single "none" action on the lower half of the graph. It is expected, as the optimal policy involves applying torque in the same direction as the car's current speed. Upon analyzing the covariance matrices of components \#3 and \#6, negative covariances were found between the state variables and the "left" action, meaning that when the car is moving left (negative speed), the Q-value for the "left" action increases, which matches the expected policy. This is something impossible to achieve with local feature representations like RBF or tile coding, since components only generalize the state space and Q-values do not vary inside a single unit. This explains why Q-IGMN can solve problems with very few Gaussian components (often 1), which also, in turn, makes it very fast.

\section{Conclusion}\label{sec:conclusion}
This work presented the combination of incremental Gaussian mixture models with reinforcement learning. IGMN was employed as a function approximator for Q-values of three classic continuous reinforcement learning tasks. Results show that it presents astounding data-efficiency, learning the three tasks within a few episodes. This can be attributed to the algorithm's similarity to double Q-Learning, delayed Q-Learning, and trace decays, all of which are known to improve data-efficiency, combined with a data-efficient function approximator itself. However, due to time and hardware constraints, the algorithm could only be tested on relatively simple environments, where this performance is not that impressive. New experiments in more complex environments (e.g., Atari) should be conducted in future works to assess the generality and scalability of this solution.

An interesting discovery found while using experience replay is that some drawbacks of conventional neural networks that require some workarounds are not present in IGMN, allowing it to take full advantage of the procedure. For instance, the IGMN does not require i.i.d. data, so experience replay does not need to be sampled randomly. Also, due to its high data-efficiency, only a single scan through the experience replay buffer is necessary. By only updating the model sporadically, simultaneous use of the Q-function estimate for action selection and updating is avoided, improving convergence, which draws parallels with double Q-learning. Thus, one of the main contributions of this research is to show how non-mainstream algorithms can be successfully combined with reinforcement learning, suggesting that neural networks (in their current forms) trained by gradient descent are not the only possibility and possibly not the best.

The main obstacles found reside in the non-stationary nature of the Q-value function, which is not appropriate for the learning rate annealing schedule of IGMN. IGMN finds means of input data, while the Q-value updates require an emphasis on more recent data, in some cases even totally overwriting previous inaccurate values (due to bootstrapping). We fixed this issue by providing a separate learning rate for the Q-values, as well as implementing a tighter control over its variance to avoid singularities.

Another obstacle found during the experiments was IGMN's high sensitivity to its hyper-parameters ($\tau$ and $\bsigma_{ini}^2$). A very small change in these parameters is enough to make the algorithm create more or less Gaussian components, which has a huge impact on the final model, like a butterfly effect. Moreover, the excessive creation of components can cause overfitting, as well as catastrophic forgetting. New components can have large overlapping regions with old components, interfering with what has been previously learned. Thus, we propose that, in future works, this issue must be addressed with a certain urgency, as this can deter practical use of the IGMN (the same issue was also found by \citep{koert2018online} and explored by \citep{chamby2018adaptive}).

\bibliographystyle{sbc}
\bibliography{bibliography.bib}

\end{document}